\documentclass{ametsocV6.1_arxiv}

\newcommand{\arrtime}{{\bm{\tau}}}
\newcommand{\arrtimegen}{{\bm{\tau}^{\bm{g}}}}
\newcommand{\arrtimeRV}{{\bm{T}}}
\newcommand{\arrtimemeas}{{\bm{\bar{\tau}}}}
\newcommand{\arrtimemeasRV}{{\bm{\bar{T}}}}
\newcommand{\cond}{{P_{\bm{T}|\bm{\bar{T}}}}}
\newcommand{\condgen}{{P_{\bm{T}|\bm{\bar{T}}}^{\bm{g}}}}
\newcommand{\joint}{{P_{\bm{T \bar{T}}}}}
\newcommand{\latent}{\bm{z}}
\newcommand{\latentdist}{P_{\bm{Z}}}

\title{Generative Algorithms for Fusion of Physics-Based Wildfire Spread Models with Satellite Data for Initializing Wildfire Forecasts}

\authors{Bryan Shaddy,\aff{a}\correspondingauthor{Bryan Shaddy, bshaddy@usc.edu \\ \emph{This Work has not yet been peer-reviewed and is provided by the contributing Author(s) as a means to ensure timely dissemination of scholarly and technical Work on a noncommercial basis. Copyright and all rights therein are maintained by the Author(s) or by other copyright owners. It is understood that all persons copying this information will adhere to the terms and constraints invoked by each Author's copyright. This Work may not be reposted without explicit permission of the copyright owner.}} 
Deep Ray,\aff{a,b} 
Angel Farguell,\aff{c} 
Valentina Calaza,\aff{a}
Jan Mandel,\aff{d} 
James Haley,\aff{e} 
Kyle Hilburn,\aff{e} 
Derek V. Mallia,\aff{f} 
Adam Kochanski,\aff{c} 
and Assad Oberai\aff{a}
}

\affiliation{\aff{a}{Department of Aerospace and Mechanical Engineering, University of Southern California, Los Angeles, California} \\
\aff{b}{Department of Mathematics, University of Maryland, College Park, Maryland}\\ 
\aff{c}{Wildfire Interdisciplinary Research Center, San Jose State University, San Jose, California}\\ 
\aff{d}{Department of Mathematical and Statistical Sciences, University of Colorado Denver, Denver, Colorado}\\
\aff{e}{Cooperative Institute for Research in the Atmosphere, Colorado State University, Fort Collins, Colorado}\\ 
\aff{f}{Department of Atmospheric Sciences, University of Utah, Salt Lake City, Utah}
}

\abstract{Increases in wildfire activity and the resulting impacts have prompted the development of high-resolution wildfire behavior models for forecasting fire spread. Recent progress in using satellites to detect fire locations further provides the opportunity to use measurements to improve fire spread forecasts from numerical models through data assimilation. This work develops a method for inferring the history of a wildfire from satellite measurements, providing the necessary information to initialize coupled atmosphere-wildfire models from a measured wildfire state in a physics-informed approach. The fire arrival time, which is the time the fire reaches a given spatial location, acts as a succinct representation of the history of a wildfire. In this work, a conditional Wasserstein Generative Adversarial Network (cWGAN), trained with WRF-SFIRE simulations, is used to infer the fire arrival time from satellite active fire data. The cWGAN is used to produce samples of likely fire arrival times from the conditional distribution of arrival times given satellite active fire detections. Samples produced by the cWGAN are further used to assess the uncertainty of predictions. The cWGAN is tested on four California wildfires occurring between 2020 and 2022, and predictions for fire extent are compared against high resolution airborne infrared measurements. Further, the predicted ignition times are compared with reported ignition times. An average Sorensen's coefficient of 0.81 for the fire perimeters and an average ignition time error of 32 minutes suggest that the method is highly accurate.}

\begin{document}

\maketitle

\section{Introduction}
Recent decades have seen increases in both wildfire frequency and severity, with parts of the western United States being some of the most impacted \citep{westerling2006warming,dennison2014large}. The increase in wildfire activity is believed to be tied to global warming, with climate predictions for North America's west coast indicating wetter winters, drier summers, and more heat in years to come, leading to conditions conducive to large wildfires \citep{williams2019observed,liu2010trends}. Wildfires impact air quality, cause damage through the destruction of property and harm to health, and negatively influence atmospheric composition \citep{jaffe2008interannual,wang2021economic,aguilera2021wildfire,solomon2022stratospheric}. With current trends expected to continue, it is important to further the advancement of wildfire prediction models to accurately forecast wildfire spread and resulting smoke dispersion. 

As climate change drives wildfires to grow larger and more intense, their interactions with the atmosphere are becoming evermore important in understanding their behavior. Large wildfires have been observed to create their own weather through strong convective updrafts resulting from the immense heat released to the atmosphere \citep{lareau2017mean,lareau2018carr}. These modifications to local meteorology then feed back into the spread of the fire through the important two-way atmosphere-wildfire coupling. The importance of wildfire-atmosphere interactions has lead to the development of a new generation of wildfire-atmosphere models which aim to capture these critical interactions \citep{bakhshaii2019review}. State of the art wildfire models have also benefited from advancements in computational power, which has made it possible to run wildfire simulations with high-resolution in operational settings \citep{kochanski2013real,jimenez2018high,mandel2019interactive}. However, issues of accumulated model errors leading to degraded forecasts still remain. Consequently, there is an interest in assimilating data from new measurement platforms into wildfire forecasts to address these issues and to allow for the use of wildfire models to accurately forecast fire spread in an operational setting. 

Remote sensing capabilities have similarly advanced in recent decades. In the context of monitoring wildfire spread there are typically two main sources of measurement data, which include airborne infrared (IR) fire extent perimeters and satellite-based active fire (AF) products. Airborne IR fire perimeters are provided through the National Infrared Operations (NIROPS) program and are generally considered to be highly spatially accurate representations of fire extent at the time of measurement \citep{greenfield2003phoenix}. However, these measurements are typically made once a day and delays in their public availability hinder their use as a basis for data assimilation in an operational setting. Measurement of IR perimeters may also be impacted by weather or resource availability. AF satellite data products on the other hand can provide the location of actively burning regions with a higher temporal frequency of around 12 hours per each low-Earth orbiting satellite, dependent on fire location. Some of the most prevalent AF data products are provided by thermal imaging sensors on board polar-orbiting satellites including the Moderate Resolution Imaging Spectroradiometer (MODIS) on board the Terra and Aqua satellites and the Visible Infrared Imaging Radiometer Suite (VIIRS) on board the Suomi-NPP and NOAA-20 satellites. These systems provide AF detections with resolutions between 375 m and 1 km. However, even these spatio-temoral resolutions are not sufficient to directly incorporate AF data in to current wildfire spread models \citep{schroeder2016visible,giglio2016collection}. Furthermore, these measurements can have artifacts from sun glint, clouds, or topography. While AF satellite data products cannot be directly used in wildfire spread models, they are a useful source for input to data assimilation algorithms which use these products and simulations from wildfire spread models to infer or initialize the state of a wildfire. 

For a dynamic coupled atmosphere-wildfire spread model, the state of the system is represented by all the wildfire and atmosphere variables in the simulation. These include wildfire variables like fire position and fuel availability, in addition to atmosphere variables such as temperature and wind speed. Further, there are two types of equally useful data assimilation procedures that can be applied to these state variables. One involves using new satellite measurements to update the state variables in an ongoing simulation \citep{mandel2014data}. The other involves using satellite measurements acquired in the initial period of the wildfire (the first 72 hours, for example) to determine the initial condition for these state variables, from which a simulation may be started \citep{farguell2021machine}. This problem of determining the initial condition of the state variables is the focus of our work. 

In \cite{mandel2012assimilation}, it was observed that if the precise history of a wildfire during its initial spread was known, then this history could be prescribed within a coupled wildfire-weather model, thereby generating the correct initial state of both the wildfire and atmosphere variables. In other words, this history could be used to spin-up the atmosphere with the right amount of heat and mass flux added at the right place and time, yielding the correct atmospheric state which is in sync with the fire state at the end of the initial phase. Thus, it was recognized that the data assimilation problem of determining the initial condition could be transformed to one of determining the history of the fire in the initial period. It was further shown that this history is succinctly represented by the fire arrival time map which contains the precise time the fire arrives at a given location \citep{mandel2012assimilation,mandel2014recent}. With these two key observations, the data assimilation problem reduces to: given satellite measurements of active fire made during the initial phase of spread, determine the high-resolution fire arrival time for this period. 

Previous methods to solve this problem have used interpolation between airborne infrared fire perimeters to create higher-resolution fire arrival times with some success \citep{kochanski2019modeling,mallia2020evaluating}. However, the coarse temporal frequency with which these measurements are made, along with delays in availability, limit the use of this approach in operational settings. Turning to the use of active fire satellite measurements for this data assimilation problem, a number of geospatial interpolation schemes have previously been used \citep{veraverbeke2014mapping,scaduto2020satellite,parks2014mapping}. Satellite active fire detection data was utilized in \cite{mandel2014data} to estimate the fire arrival time by penalizing the difference between the prior obtained from model results and satellite active fire detection pixels, however this approach did not take the rate of spread into account. It was then considered to minimize the residual of a differential equation model of fire propagation (the eikonal equation $\|\nabla{u}\| = \frac{1}{R}$, where $R=R(u,x,y)$ is the rate of spread and $u$ is the fire arrival time) subject to constraints derived from data, however this approach faced numerical difficulties \citep{farguell2018assimilation}. To overcome previous difficulties, in \cite{farguell2021machine} a machine learning based method for estimating fire arrival times was developed. This method uses a support vector machine (SVM) to estimate the fire arrival time based on satellite active fire and clear ground detection pixels. The SVM method works by finding an optimal separation between the fire and no-fire locations in space and time to construct the fire arrival time. While the SVM method has provided good results, it does not incorporate any of the physics inherent to wildfire spread into estimates, and further does not provide information about uncertainty, which is desirable when relying on error prone satellite measurements. The method described in this work addresses both of these limitations. 

The point of departure of the method presented in this manuscript is a probabilistic interpretation of the problem. We treat both the measured active fire pixels and the desired fire arrival time field as random vectors. The inference problem we wish to solve is one of quantifying the conditional probability distribution for the fire arrival time conditioned on a given measurement of the active fire pixels. We recognize that the measured fire pixels and the arrival time field are both high-dimensional random vectors, which makes this problem challenging to solve. To address this challenge, we utilize a conditional generative algorithm called the conditional Wasserstein Generative Adversarial Network (cWGAN). A cWGAN relies on the expressivity of a deep neural network and the concept of adversarial loss to learn and then sample from a conditional probability distribution. Its training requires the use of samples from the joint probability distribution of the field to be inferred (fire arrival time) and the measurements (active fire pixels). We generate these by employing WRF-SFIRE to produce physically consistent fire arrival time maps, to which we then apply an approximation of the measurement operator. This approach does not require any satellite measurement data to train the network, and allows us to inject the appropriate physics into the inference problem. Once trained, we apply the algorithm retrospectively to four recent wildfires in California and assess its performance by comparing its predictions with IR fire extent perimeters and reported ignition times, which are treated as ground truth. We also include predictions from the SVM based algorithm for comparison and describe the relative benefits of the two approaches. 

The remainder of this manuscript is organized as follows. In Section \ref{prob_fomulation}, we provide a mathematical formulation for the data assimilation problem. In Section \ref{methods}, we describe the cWGAN algorithm used to solve this problem. In Section \ref{results}, we apply this algorithm to wildfires and quantify its performance. We end with conclusions and remarks for future work in Section \ref{conclusion}.

\section{Problem formulation} \label{prob_fomulation}
We let $\arrtime$ denote the matrix of fire arrival times whose components $\tau_{ij}$ represent the fire arrival times for the $i$-th pixel along the longitude and the $j$-th pixel along the latitude. We assume that there are $N_\tau$ such pixels and therefore $\arrtime \in \Omega_{\arrtime} \subset \mathbb{R}^{N_\tau} $. The size of each pixel is $60 \times 60 $ meters. 

A measurement operator $M$ may be applied which transforms $\arrtime$ into a coarse, sparse, and noisy measurement $\arrtimemeas$. That is to say, we may use the mapping $M: \Omega_{\arrtime} \rightarrow \Omega_{\arrtimemeas}$, which takes as input the complete and smooth fire arrival time and produces the corresponding measurement $\arrtimemeas \in \Omega_{\arrtimemeas}$. Since the arrival time and the measurement are defined on the same grid, $\Omega_{\arrtimemeas} = \Omega_{\arrtime}$. We note that the measurement operator $M$ may easily be approximated to produce the mapping from fire arrival time to active fires satellite measurements (see Section~\ref{methods}\ref{training} for the precise definition); however here we are interested in the inverse problem, which maps from $\arrtimemeas$ to $\arrtime$ and is much more challenging to solve. 

We recognize that a single measurement can correspond to a distribution of likely fire arrival times, and to cope with the ill-posed nature of this problem we adopt a probabilistic approach. We let the inferred field $\arrtime$ and the measurements $\arrtimemeas$ be modeled by random variables $\arrtimeRV$ and $\arrtimemeasRV$, respectively. Following this, we recognize that given a measurement $\arrtimemeas$, we are interested in learning and generating samples from the conditional distribution $\cond$. We accomplish this through the following steps:
\begin{enumerate}
    \item Generate $N$ pairwise samples of arrival times and measurements $(\arrtime^{(i)},\arrtimemeas^{(i)}), \; i = 1, \cdots, N$ sampled from the joint distribution $\joint$. This is accomplished by using WRF-SFIRE (and data augmentation)to generate $N$ instances of arrival times $\arrtime^{(i)}$ and then applying the measurement operator to obtain the corresponding $\arrtimemeas^{(i)}$. 
    \item Use this data to train the generator and critic sub-networks of a conditional Wasserstein Generative Adversarial Network (cWGAN).  
    \item Use active fire satellite measurements of a wildfire as input to the trained generator to produce samples of the arrival time from the conditional distribution. Further, use these samples to generate statistics of interest, which include the pixel-wise mean and standard deviation in arrival time. 
\end{enumerate}

In the following Section we provide a brief summary of the cWGAN. The interested reader is referred to \cite{adler2018deep, ray2022efficacy, ray2023solution} for further details. Thereafter, in Section~\ref{methods}\ref{training} we describe the generation of the training data for the cWGAN and the training procedure. Finally in Section~\ref{results} we describe the application of the trained cWGAN to four recent wildfires in California.

\section{Conditional Wasserstein Generative Adversarial Networks (cWGANs)} \label{methods}
The cWGAN consists of two subnetworks, a generator $\bm{g}$ and a critic $d$. The generator $\bm{g}$ is given by the mapping $\bm{g}: \Omega_{\latent} \times \Omega_{\arrtimemeas} \rightarrow \Omega_{\arrtime}$, where $\latent \in \Omega_{\latent} \subset \mathbb{R}^{N_z}$ is a latent variable modeled using the random variable $\bm{Z}$ with distribution $\latentdist$. The distribution $\latentdist$ is selected such that it is easy to sample from, such as a multivariate Gaussian distribution. The critic $d$ is given by the mapping $d: \Omega_{\arrtime} \times \Omega_{\arrtimemeas} \rightarrow \mathbb{R}$.

For a given measurement $\arrtimemeas$, the generator $\bm{g}$ produces samples $\arrtimegen=\bm{g}(\latent,\arrtimemeas), \; \latent \sim \latentdist$ from the learned conditional distribution $\condgen (\arrtime| \arrtimemeas)$. The training of the cWAN requires this distribution to be close to the true conditional distribution $\cond (\arrtime | \arrtimemeas)$ in the Wasserstein-1 metric. The cWGAN is trained using the following objective function, 
\begin{equation} \label{obj_fun}
    \mathcal{L}(d,\bm{g}) = \mathop{\mathbb{E}}_{\substack{(\arrtime,\arrtimemeas) \sim \joint}} [d(\arrtime,\arrtimemeas)] - \mathop{\mathbb{E}}_{\substack{\arrtimegen \sim \condgen\\ \arrtimemeas \sim P_{\arrtimemeasRV}}} [d(\arrtimegen,\arrtimemeas)].
\end{equation}
The optimal generator and critic ($\bm{g}^*$ and $d^*$, respectively) are determined by solving the min-max problem 
\begin{equation} \label{minmax}
    (d^*,\bm{g}^*) = \mathop{\mathrm{arg \, min}}_{\bm{g}} \; \mathop{\mathrm{arg \, max}}_{d} \; \mathcal{L}(d,\bm{g}).
\end{equation}
Assuming the critic $d$ is 1-Lipschitz in both its arguments, it is shown in \cite{ray2023solution} that the $\bm{g}^*$, found by solving the min-max problem given by Eq.~\eqref{minmax}, can be used to approximate the true conditional distribution. More precisely, for any continuous, bounded function $\ell (\arrtime)$ defined on $\Omega_{\arrtime}$, and $\epsilon > 0$, we may select a generator with a sufficiently large number of learnable parameters such that 
\begin{equation} \label{conv}
    |\mathop{\mathbb{E}}_{\arrtime \sim \cond} [\ell(\arrtime)]  - \mathop{\mathbb{E}}_{\arrtime \sim P_{\bm{T}|\bm{\bar{T}}}^{\bm{g}^*}} [\ell(\arrtime)] | < \epsilon. 
\end{equation}

This implies that once the generator has been trained it may be used to approximate the statistics from the true conditional distribution. In particular,
\begin{equation} \label{stats}
    \mathop{\mathbb{E}}_{\arrtime \sim \cond} [\ell(\arrtime)] \approx \frac{1}{K} \sum\limits_{i=1}^K \ell (\bm{g}^*(\latent^{(i)},\arrtimemeas)), \quad \latent^{(i)} \sim \latentdist.
\end{equation}
Following Eq.~\eqref{stats}, the pixel-wise mean prediction for $\arrtime$ based on a given $\arrtimemeas$ can be computed by setting $\ell(\arrtime) = \arrtime$. Similarly, the pixel-wise variance may be computed by setting $\ell(\arrtime)=(\arrtime-\mathbb{E}[\arrtime])^2$, which proves useful for quantifying the uncertainty in our prediction. 

\subsection{cWGAN architecture} \label{architecture}
\begin{figure}
    \centering
    \includegraphics[width=\linewidth]{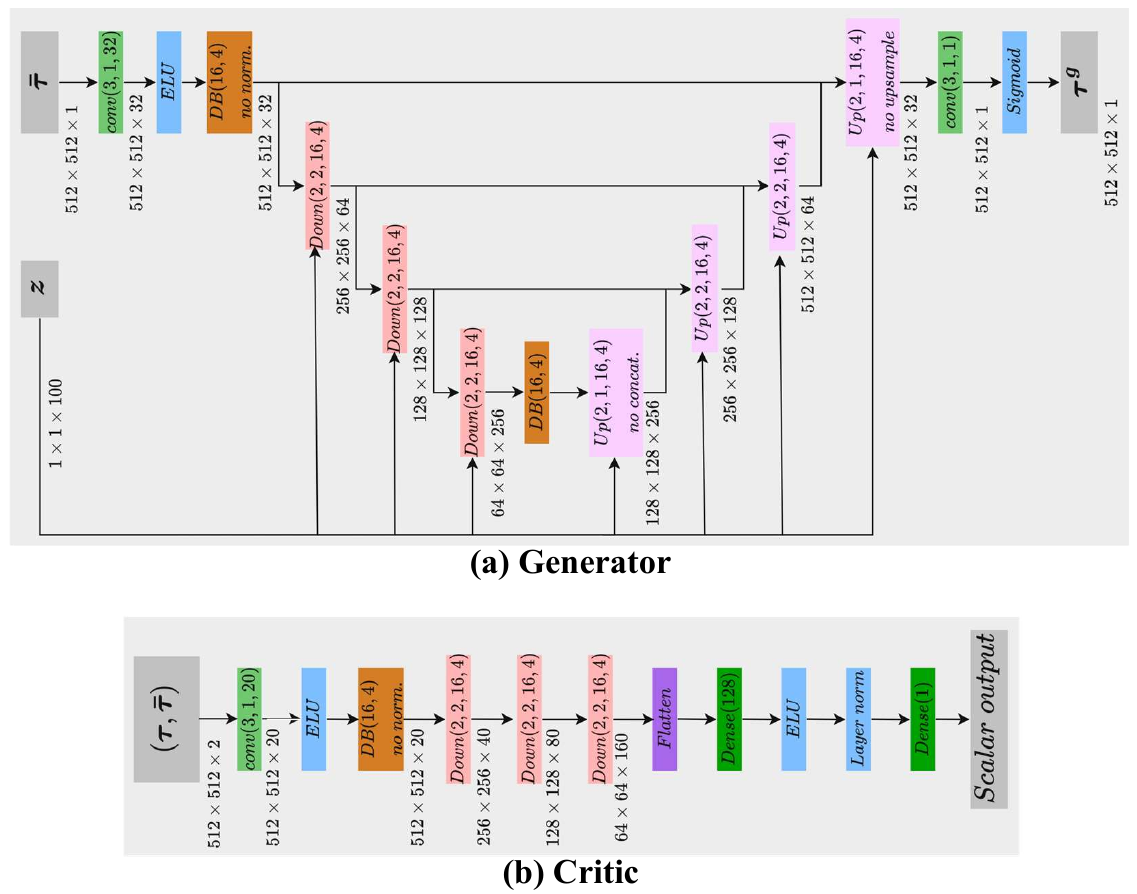}
    \caption{Architecture of (a) generator and (b) critic used in cWGAN. The $Down$, $Up$ and Dense ($DB$) blocks are described in Fig.~\ref{fig:db_down_up}.}
    \label{fig:cWGAN_schematic}
\end{figure}

The architectures of the generator and critic used here are based on \cite{ray2022efficacy} and are shown in Fig.~\ref{fig:cWGAN_schematic}. The generator $\bm{g}$ has a U-Net architecture which takes as input the measurement $\arrtimemeas$ and the latent vector $\latent$. The latent information is injected into the U-Net at various scales using conditional instance normalization (CIN) \citep{dumoulin2016learned}. This allows the latent dimension $N_{z}$ to be selected independent of $N_{\tau}$, and introduces stochasticity at multiple scales of the U-Net, thereby overcoming the problem of mode collapse which previously required the use of more complicated critic architectures \citep{ray2022efficacy,adler2018deep}.

In the U-Net the residual blocks used in \cite{ray2022efficacy} are replaced with dense blocks, which have been shown to lead to superior performance while reducing the number of trainable parameters \citep{huang2017densely}. Dense blocks are denoted as DB($k,n$), where $n$ is the number of sub-blocks in the dense block and $k$ is the number of output features in all but the last sub-block of the dense block. The generator architecture also contains down-sampling blocks denoted by Down($p,q,k,n$), which coarsen the resolution of inputs by a factor of $p$ and increase the number of channels by a factor of $q$. The $k$ and $n$ listed for a down-sampling block refer to the parameters of the dense block contained as a sub-block of the down-sampling block. Up-sampling blocks are denoted as Up($p,q,k,n$), where $p$ is the factor by which resolution is refined and $q$ is the factor by which the number of channels decreases. Again, $k$ and $n$ refer to the dense-block included in the up-sampling block. Values for $k$, $n$, $p$, and $q$ used in this work are included in the schematic shown in Fig.~\ref{fig:cWGAN_schematic}.

\begin{figure}
    \centering
    \includegraphics[width=\linewidth]{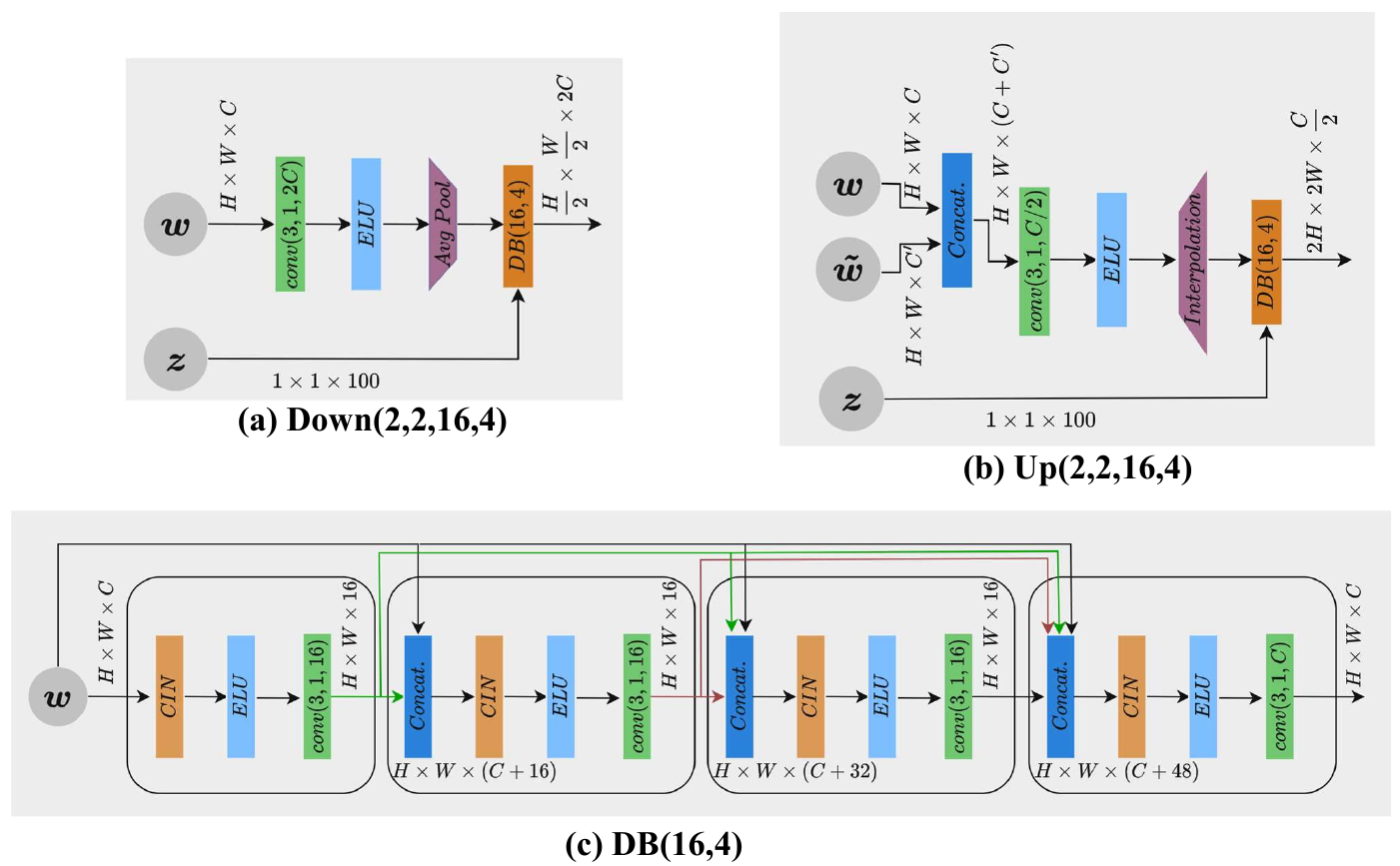}
    \caption{Architecture of (a) down-sample block, (b) up-sample block, and (c) dense block, with the values $p=2$, $q=2$, $k=16$, and $n=4$ used for this work shown. }
    \label{fig:db_down_up}
\end{figure}

The architecture of the critic follows that of \cite{ray2022efficacy}, with a relatively simple architecture comprising dense blocks and down-sampling blocks, followed by fully connected layers which result in a scalar output. Schematics for the dense blocks, down-sample blocks, and up-sample blocks used here are provided in Fig.~\ref{fig:db_down_up}, with the corresponding network parameters used for this work, $p=2$, $q=2$, $k=16$, and $n=4$, also shown.

\subsection{Training the cWGAN} \label{training}
To train the cWGAN, we begin by drawing samples of $\arrtime$ from the prior marginal distribution $P_{\arrtimeRV}$ by performing simulations with the coupled atmosphere-wildfire model WRF-SFIRE. To each instance of $\arrtime$ an approximation of the measurement operator $M$ is applied to produce a corresponding measurement $\arrtimemeas=M(\arrtime)$, producing samples from the conditional density $P_{\arrtimemeasRV|\arrtimeRV}$. This yields the pairs $(\arrtime^{(i)},\arrtimemeas^{(i)})$ from the joint distribution $\joint$ which are used to train cWGAN. 

The fire arrival times are computed from 20 idealized WRF-SFIRE simulations, each considering 2 day fire spread over flat terrain with a uniform fuel type of brush. The initial condition used for wind in all simulations consists of a logarithmic profile up to 2 km and constant wind speed above 2 km, with wind prescribed uniformly in one direction. The initial wind profile was maintained and the wind magnitude 10 m from the surface was varied randomly from a uniform distribution between $0-5 \; \mathrm{m \; s^{-1}}$ to produce the 20 simulation results. 

The simulations used an atmospheric mesh of size $128 \times 128$, with a resolution of 300 m, giving a domain size of $38.4 \times 38.4$ km. The fire mesh was refined from the atmospheric mesh by a factor of 10, for a mesh size of $1280 \times 1280$ and a resolution of 30 m, with point ignitions located in the center of the domain. The highest altitude for the atmospheric model was fixed at 4500 m, and the model considered 41 vertical levels, open boundary conditions, and a fixed fuel moisture content of 18\%. 

The fire arrival time results from the 20 WRF-SFIRE simulations were cropped to a new grid size of $1024 \times 1024$, on a domain of size $30.72 \times 30.72$ km. The fire arrival times were then coarsened from their initial resolution of 30 m to a resolution of 60 m, giving a final grid size of $512 \times 512$. Cropping and coarsening was done to reduce the size of the input to the cWGAN, and consequently reduce the size of network needed for this work. Data augmentation was then performed to increase the total number of fire arrival time maps available for training. Augmentation was done by rotating the fire arrival time maps randomly between 0 and 360 degrees about their center, and then translating them within a box of size $9 \times 9$ km located at the center of the domain. Each WRF-SFIRE arrival time lead to 500 augmented samples thereby generating a total of 10,000 fire arrival times $\arrtime^{(i)}$. The augmentation of the fire arrival times was enabled by the rotational and translational symmetry of the simulations, which consisted of flat terrain, uniform fuel, uniform initial wind direction, and point ignition. 

An approximation to the measurement operator $M$ was applied to the samples of $\arrtime^{(i)}$ to generate the corresponding measurement $\arrtimemeas^{(i)}$. The measurement operator was constructed to replicate the high resolution (375 m) VIIRS L2 AF data in the following steps:
\begin{enumerate}
    \item Coarsen $\arrtime^{(i)}$ to a resolution of 375 m using nearest neighbor interpolation.
    \item Select four measurement times ($t_i, \; i = 1,\cdots, 4$) from a uniform distribution between 2 hours and 48 hours and sort them in ascending order. 
    \item For each measurement time, $t_i$, create a time interval $(t_i - \delta^{(-)}, t_i)$ where $\delta^{(-)}$ is selected from $\mathcal{U}(6,12)$, where $\mathcal{U}$ denotes the uniform probability distribution. If $t_i - \delta^{(-)} < 0$, set it to 0.
    \item Create four copies of the coarsened $\arrtime^{(i)}$, one for each time interval, and denote them by $\arrtime^{(i)}_j, \; j = 1, \cdots, 4$. 
    \item To each $\arrtime^{(i)}_j$, apply a distinct knowledge mask that randomly eliminates 50\% of the fire arrival time values. Set eliminated pixels to a background value.   
    \item For each $\arrtime^{(i)}_j$, set fire arrival time pixels falling within the associated time interval $(t_i - \delta^{(-)}, t_i)$ to $t_i$. Set the remainder to a background value, to be assigned later.
    \item Combine the four measurements into a single consolidated measurement by selecting $\arrtime^{(i)} = \min_j(\arrtime^{(i)}_j)$ for each pixel.   
    \item Eliminate three $3 \times 3$ km patches with locations selected at random. Set the values in these patches to the background value to emulate measurement obstruction. 
    \item Resample $\arrtime^{(i)}$ back to the original size of $512 \times 512$ pixels.
    \item Add $\delta \in \mathcal{U}(0,24)$ hours to the arrival time and measurement pair $(\arrtime^{(i)},\arrtimemeas^{(i)})$ to account for the fact that the ignition time is typically unknown. 
    \item Normalize the arrival time and measurement pair $(\arrtime^{(i)},\arrtimemeas^{(i)})$ to be in the interval $[0,1]$ by dividing it by $72$ hours and setting the background value to 1. 
\end{enumerate}
 
\begin{figure}
    \centering
    \includegraphics[width=\linewidth]{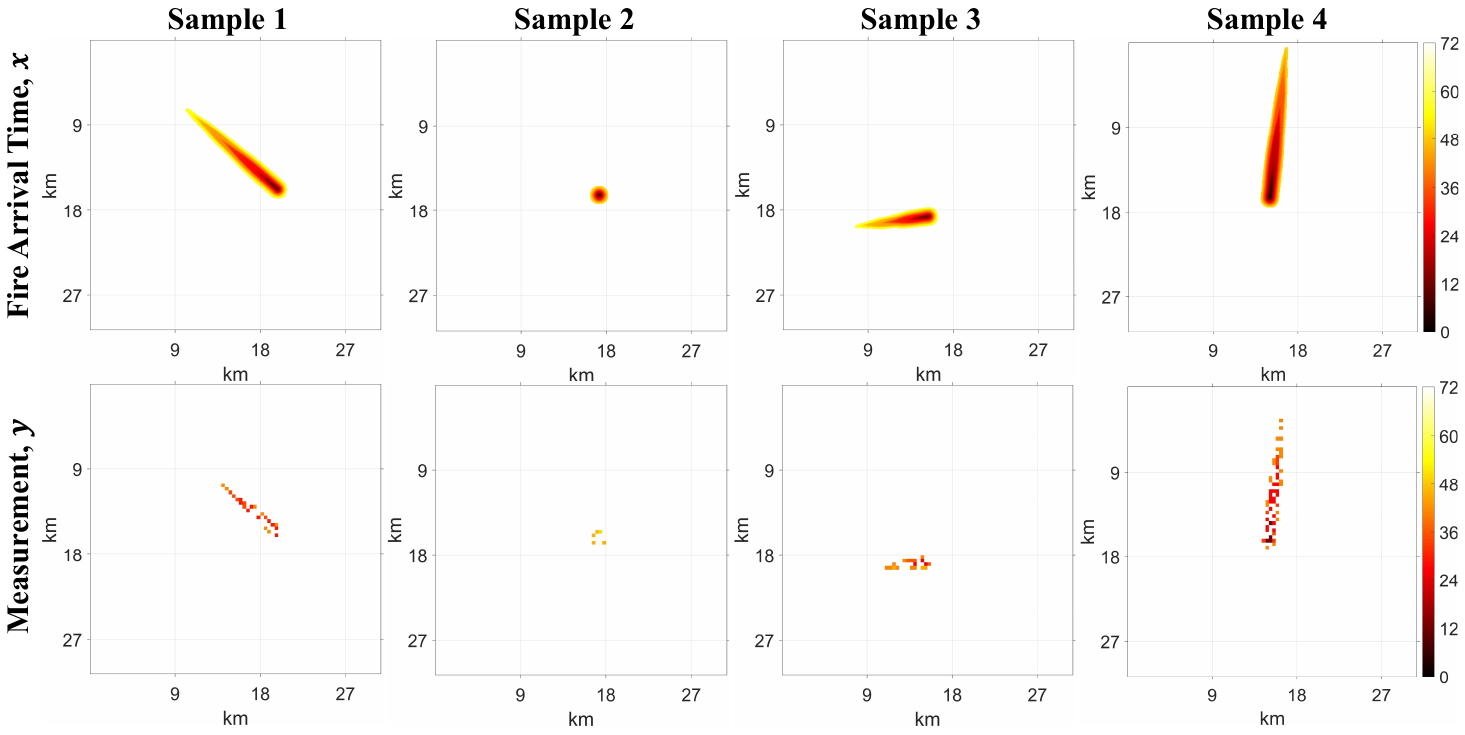}
    \caption{Sample data pairs from the training set, with true fire arrival times $\arrtime$ in the first row and corresponding measurements $\arrtimemeas$ in the second row. Here fire arrival time values represent hours from the start of the day on which ignition occurred.}
    \label{fig:training_samples}
\end{figure}

The data set was split so 8,000 samples were used for training and 2,000 samples were reserved for validation (selecting optimal hyperparameters). Sample data pairs from the training data set are shown in Fig.~\ref{fig:training_samples}. 

The training of the cWGAN was tracked using a mismatch term defined as the 2-norm of the difference between the generated fire arrival time $\arrtimegen=\bm{g}(\latent,\arrtimemeas)$ and the true fire arrival time $\arrtime$ from the training set. Additional metrics tracked included the approximate Wasserstein-1 distance measured as the difference between the scalar critic outputs for the true and generated input values $\arrtime$ and $\arrtimegen$, and the critic and generator losses. The cWGAN was trained for 200 epochs, following which the mismatch term had reached an acceptably low value, while also balancing consideration of the training time.

\section{Results} \label{results}
In this section we validate the cWGAN fire arrival time predictions for four retrospective California wildfires. Comparisons are made against high resolution fire extent perimeters and reported ignition times. Further comparisons are made with the SVM method described in \cite{farguell2021machine}.

\subsection{Test cases}
The criteria for selecting wildfires for validation are:
\begin{enumerate}
    \item Sufficient VIIRS AF measurements (for input) and at least one fire extent measurement (for validation), both within the first 48 hours of the wildfire ignition.
    \item Wildfires whose extent does not exceed the domain size considered in the training data.
\end{enumerate}

Following these considerations, four fires are selected for validation: Bobcat, Tennant, Oak, and Mineral. These fires occurred in California between the years 2020 and 2022. It is worth noting that while the training data for the cWGAN considered flat terrain only, this was not a requirement for the fires selected for validation. Additional information about these fires is presented in Table~\ref{t1:cases}. 

\begin{table}[t]
\caption{Wildfire test cases examined using cWGAN approach. For each fire, start date (UTC), approximate ignition time (UTC), approximate initial location, fire perimeter day (UTC), and fire perimeter time (UTC) are listed. Ignition times for Tennant, Oak, and Mineral are gathered from CALFIRE reports. Ignition time for Bobcat is unavailable from CALFIRE and has been gathered from news reports, and is not used for evaluating performance.}\label{t1:cases}
\begin{center}
\begin{tabular}{cccccc}
\hline
Wildfire & Start Date & Approximate Ignition Time & Fire Perimeter Date & Fire Perimeter Time & Approximate Coordinates\\
\hline
Bobcat & 6 September 2020 & ~1900 & 8 September 2020 & 0815 & [34.26,-117.96] \\
Tennant & 28 June 2021 & 2307 & 30 June 2021 & 0605 & [41.67,-122.05] \\
Oak & 22 July 2022 & 2110 & 24 July 2022 & 0546 & [37.55,-119.92] \\
Mineral & 13 July 2020 & 2340 & 15 July 2020 & 0315 & [36.18, -120.56] \\
\hline
\end{tabular}
\end{center}
\end{table}

\subsection{Active fire satellite measurements}
The 375 m Level-2 (L2) Active Fire product from the Visible Infrared Imaging Radiometer Suite (VIIRS) on board the Suomi-NPP satellite is utilized. This data product, referred to as VNP14IMG, provides day and night fire detections globally using algorithms based on the baseline MODIS product, Thermal Anomalies and Fire. Data comes from roughly 6 minute orbital segments created from multiple scans using input data from all five 375 m I-channels (I1-I5) and the dual-gain 750 m mid-infrared M13 channel of the VIIRS system \citep{schroeder2016visible}. Detections are provided approximately $2-4$ times a day from the Suomi-NPP satellite, dependent on geographic position, with locations of active fire detections provided as latitude and longitude coordinates. Confidence information is further provided for each detection, including low, nominal, and high confidence labels. The VNP14IMG data used for this work was collected from the Level-1 and Atmosphere Archive and Distribution System Distributed Active Archive Center (LAADS DAAC) hosted by NASA. 

To preprocess VIIRS 375 m L2 AF satellite data to be used as input to the cWGAN, a domain of interest with size $30.72 \times 30.72$ km which is approximately centered on a desired wildfire is selected. The domain is discretized based on latitude and longitude coordinates and cells corresponding to AF detection locations are assigned a value based on the measurement time. Values assigned to activated cells are the number of hours since the start of the day on which ignition occurred. The measurements are then normalized using a value of 72 hours, following which remaining cells are assigned a background value of 1, putting the measurements in the range $[0,1]$, following the format of the training data.  

Measurements are further separated based on confidence level such that one version contained only high confidence detections and another version contained both high and nominal confidence detections. This results in two measurements $\arrtimemeas$ per fire, on which predictions may be conditioned. This was done for all measurements available within the first 48 hours of a fire, being sure to assign the earliest available measurement time for cells that correspond to detections in more than one satellite measurement.

\begin{figure}
    \centering
    \includegraphics[width=\linewidth]{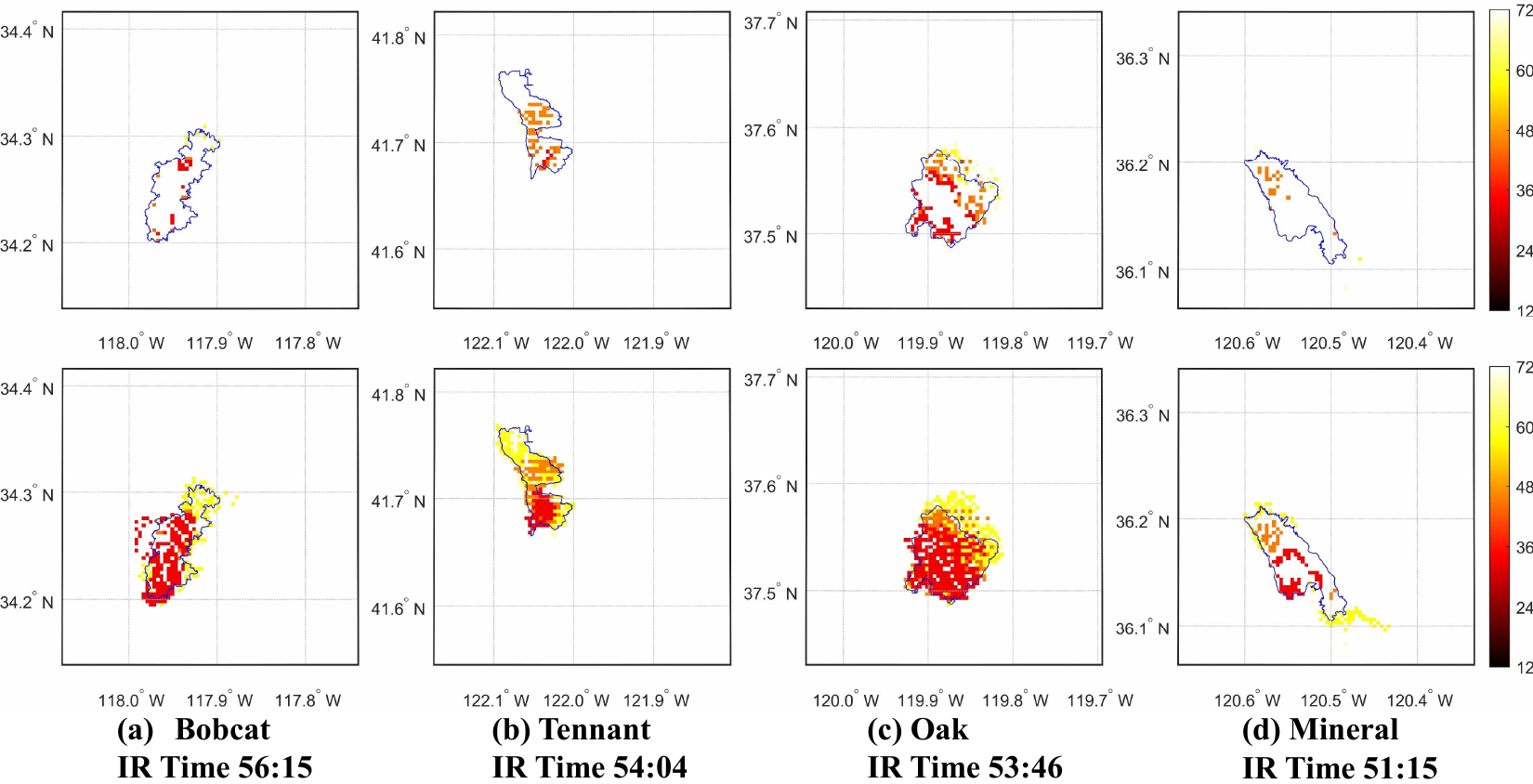}
    \caption{Measurements $\arrtimemeas$ after preprocessing of VIIRS 375 m L2 AF data for the Bobcat, Tennant, Oak, and Mineral fires, in left to right order. The first row contains high confidence detections only and the second row contains high and nominal confidence detections. AF detection colors indicate the measurement time, taken as the number of hours after the start of the ignition day. IR fire extent perimeters are additionally included, with measurement times listed in plot labels in HH:MM format, again as the number of hours after the start of ignition day. All measurements are geolocated, with longitude and latitude indicated.}
    \label{fig:meas_plots}
\end{figure}

To evaluate available AF data for each of the four test cases, the preprocessed and geolocated measurements $\arrtimemeas$ are examined relative to available IR fire extent perimeters. Shown in Fig.~\ref{fig:meas_plots} are two measurements $\arrtimemeas$ per fire, corresponding to the two confidence intervals. For each fire, one fire extent measurement made within the first 48 hours after ignition is overlaid on the measurement images for comparison. The times of the perimeter measurements are indicated in the plot labels of Fig.~\ref{fig:meas_plots} in hours and minutes (HH:MM format) from the start of ignition day. 

\subsection{Arrival time predictions and statistics} \label{sample_gen}
The two measurements (high and high + nominal confidence) for each fire are used as input to the trained cWGAN. For each measurement 200 realizations of the arrival time are generated by sampling the latent vector $\latent$ from its distribution. These realizations are combined with different weights (0.2 for the high confidence and 0.8 for the high + nominal confidence) to compute the pixel-wise mean and standard deviation plots of arrival times that are shown in Fig.~\ref{fig:prediction_stats}. There are several interesting observations to be made:

\begin{figure}
    \centering
    \includegraphics[width=\linewidth]{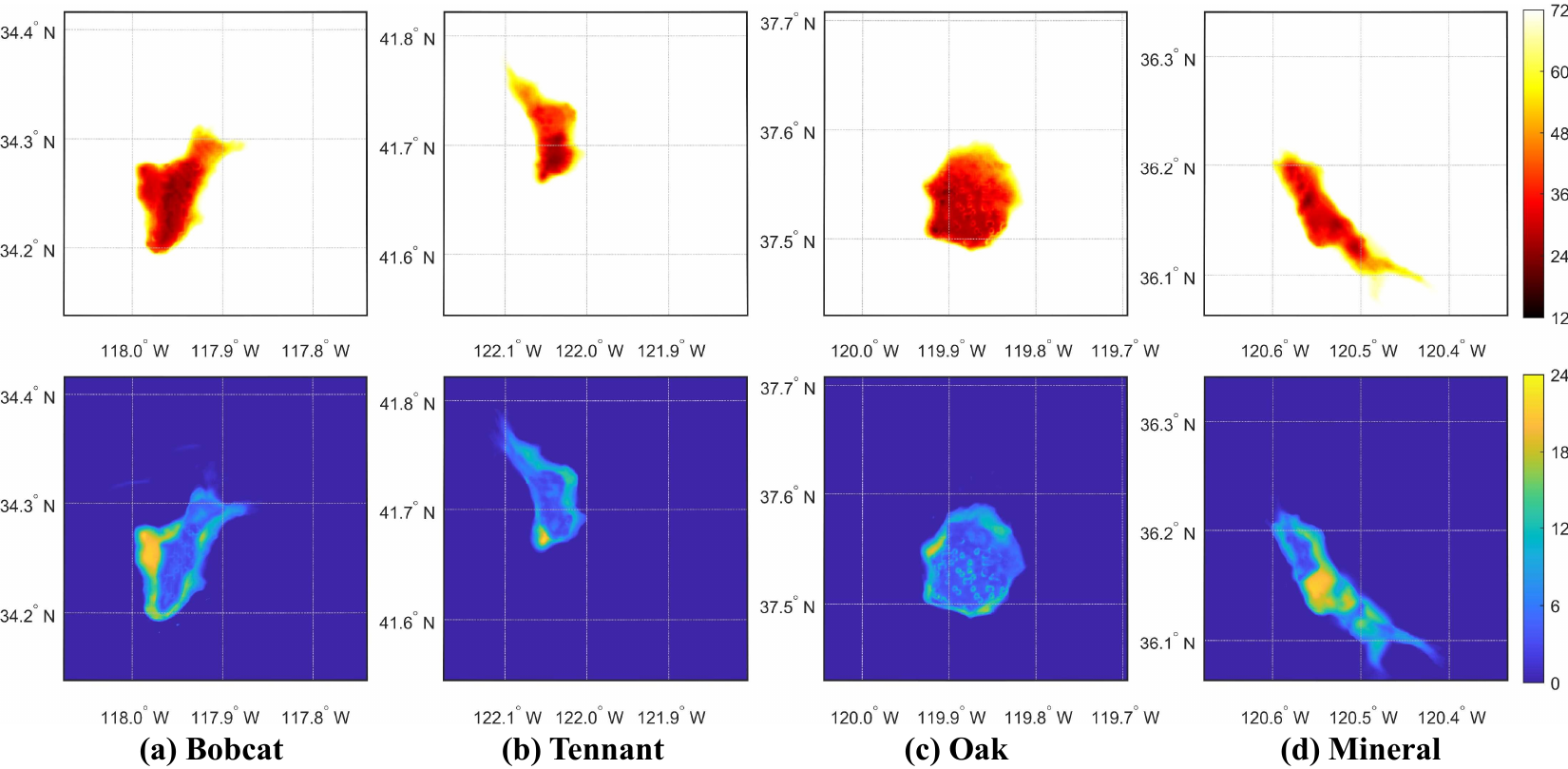}
    \caption{Weighted mean (first row) and standard deviation (second row) of fire arrival time predictions using the cWGAN approach.}
    \label{fig:prediction_stats}
\end{figure}

\begin{itemize}
    \item The mean arrival time plots appear to smooth interpolations of the measurements shown in Fig.~\ref{fig:meas_plots}. They can be used to generate a smooth sequence of fire perimeters to initialize the state variables in a coupled weather/wildfire code like WRF-SFIRE. 
    \item The standard deviation plots in Fig.~\ref{fig:prediction_stats} provide a measure of uncertainty in the predictions. We observe that in some cases and in some select locations this uncertainty is as high as 20 hours, which implies that the ensemble of predictions used to generate the mean have significant differences in these regions. 
    \item A closer look reveals that the regions of largest standard deviations are correlated with regions with significant differences between the high and high+nominal confidence measurements. This is the case for the western regions of the Bobcat fire, the southern-most tip of the Tenant fire, and the southern region of the Mineral fire. 
\end{itemize}

\begin{figure}
    \centering
    \includegraphics[width=\linewidth]{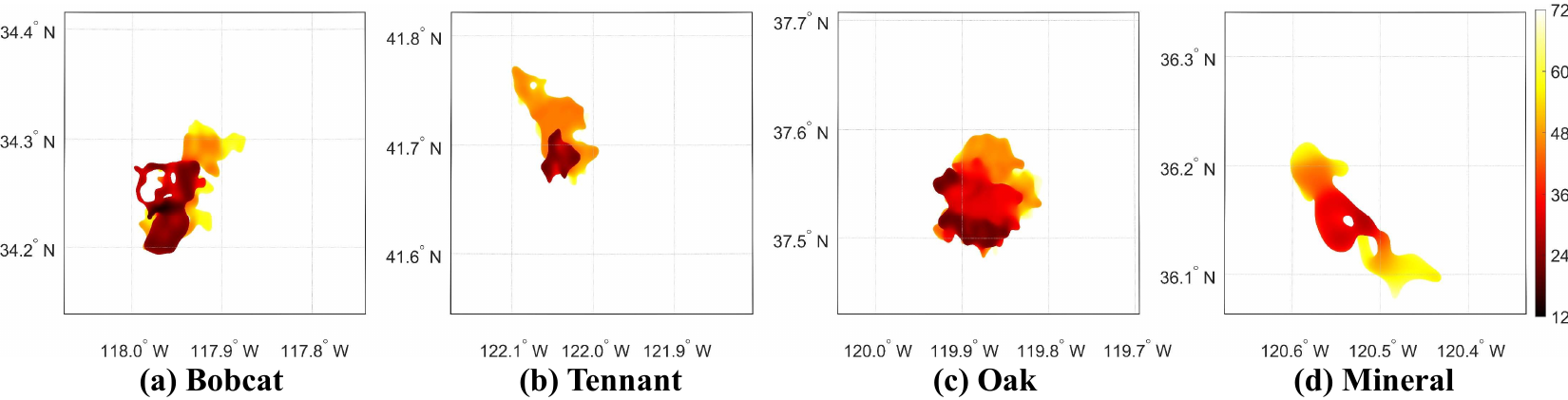}
    \caption{Fire arrival time predictions produced using the SVM method \citep{farguell2021machine}.}
    \label{fig:SVM_predictions}
\end{figure}

It is also informative to compare the fire arrival time predictions made by the cWGAN method to those produced by the SVM method described in \cite{farguell2021machine}, which is also designed to predict fire arrival times from AF satellite data. The results from the SVM method, using the same set of 375 m VIIRS L2 AF data from the Suomi-NPP satellite, are shown in Fig.~\ref{fig:SVM_predictions}. We observe that the SVM method produces islands of unburnt regions on the interior of predicted fire extents (see the Bobcat and Mineral fires), while these are absent from the cWGAN predictions. Additionally, it is noted that the SVM method does not provide any measure of uncertainty with its predictions, unlike the cWGAN approach.

\subsection{Spatial agreement with IR fire extent perimeters}
To quantitatively assess the fire arrival time predictions made by the cWGAN and SVM methods, high resolution infrared (IR) wildfire extent perimeters provided by the National Infrared Operations (NIROPS) program are used as ground-truth \citep{greenfield2003phoenix}. These perimeters are obtained from infrared sensors on-board aircraft that fly over and survey large wildfires during the night. They are considered a very accurate representation of wildfire extent at the time of measurement with accuracy on the order of meters (approximately 6.3 m/pixel). 

\begin{figure}
    \centering
    \includegraphics[width=\linewidth]{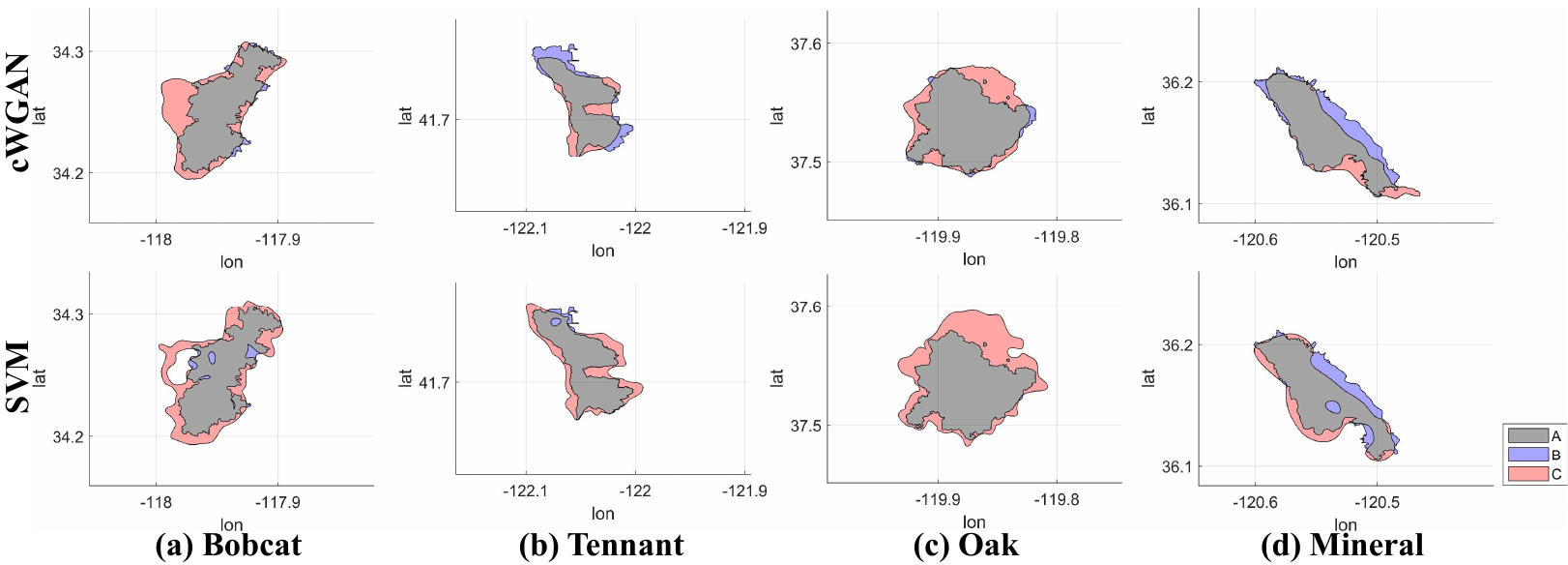}
    \caption{Plots comparing the predicted and measured IR fire extent perimeters. Grey pixels (labeled A) represent the true positive pixels, blue pixels (labeled B) represent false negative pixels, and red pixels (labeled C) represent false positive pixels.}
    \label{fig:area_discrepancy}
\end{figure}

Using the fire arrival times produced for each fire by the cWGAN and the SVM methods, along with the geolocation information used when preprocessing the AF measurements, a geolocated fire perimeter is computed for any time within 72 hours from the start of ignition day simply by plotting a contour of the fire arrival time at the prescribed time. The predicted perimeter is compared with the measured perimeter (see Fig.~\ref{fig:area_discrepancy}) by identifying the true positive pixels (burnt in both the prediction and truth), the false negative pixels (not burnt in the prediction but burnt in truth) and the false positive pixels (burnt in the prediction but not burnt in truth). These regions are marked as A, B and C, respectively, in the figure. Using these regions, three dimensionless numbers that quantify the accuracy of a prediction are computed. These are the S\o rensen–Dice coefficient (SC), the Probability of Detection (POD), and the False Alarm Ration (FAR). They are defined as
\begin{equation}
    SC = \frac{2A}{2A+B+C} \quad, \quad POD = \frac{A}{A+B} \quad, \quad FAR = \frac{C}{A+C}. 
\end{equation}
All these coefficients attain values between 0 and 1. For the SC and POD the best model yields a value of 1, whereas for the FAR a value of 0 is ideal.

Table~\ref{t2:coefficients} contains the SC, POD, and FAR values computed for the Bobcat, Tennant, Oak, and Mineral fires based on predictions by the cWGAN and SVM methods.
In all cases except the Tennant fire, the SC for the cWGAN method is higher than the SVM method. In the case of the Tennant fire, while the SC for the cWGAN prediction is lower than that for the SVM prediction, the values are close. From the POD values we conclude that the SVM method performs better, indicating that the cWGAN is less likely than the SVM method to capture the full extent of the fire. On the other hand, the FAR values for the cWGAN are better than those for the SVM method in every case, indicating the cWGAN is less likely than the SVM method to suffer from false positive errors.

\begin{table}[t]
\caption{S\o rensen's coefficient (SC), probability of detection (POD), and false alarm ratio (FAR) values obtained for the cWGAN and SVM predictions.}\label{t2:coefficients}
\begin{center}
\begin{tabular}{c|cc|cc|cc}
\hline
Wildfire & cWGAN SC & SVM SC & cWGAN POD & SVM POD & cWGAN FAR & SVM FAR \\
\hline
Bobcat & $\bm{0.80}$ & 0.77 & $\bm{0.97}$ & 0.95 & $\bm{0.32}$ & 0.35 \\
Tennant & 0.78 & $\bm{0.80}$ & 0.78 & $\bm{0.95}$ & $\bm{0.21}$ & 0.31 \\
Oak & $\bm{0.84}$ & 0.77 & 0.97 & $\bm{>0.99}$ & $\bm{0.26}$ & 0.38 \\
Mineral & $\bm{0.81}$ & 0.80 & 0.76 & $\bm{0.79}$ & $\bm{0.14}$ & 0.19 \\
\hline
\end{tabular}
\end{center}
\end{table}

\subsection{Ignition time predictions}
Both the cWGAN and SVM methods predict fire arrival times with reference to the start of the ignition day. Thus, implicitly they also generate an estimate of the ignition time for a fire, which corresponds to smallest fire arrival time in the prediction. These predictions are reported in Table~\ref{t3:ignition_times} for the Tennant, Oak, and Mineral fires, for which the ignition times published by CAL FIRE are available for comparison. In each case, the cWGAN method is significantly more accurate than the SVM method. The latter appears to bias the prediction towards the time when the first satellite measurement is available, leading to an overestimate of the ignition time. 

\begin{table}[t]
\caption{Estimated ignition times for the cWGAN and SVM predictions. Ignition times are reported in hours and minutes (HH:MM) from the start of the true ignition day. Results are given for the Tennant, Oak, and Mineral fires based on ignition times reported by CAL FIRE. The Bobcat fire is excluded as its ignition time is not reported by CAL FIRE.}\label{t3:ignition_times}
\begin{center}
\begin{tabular}{c|c|cc|cc}
\hline
Wildfire & Reported Ignition Time & cWGAN Ignition Time & SVM Ignition Time & cWGAN Error & SVM Error \\
\hline
Tennant & 23:07 & $\bm{23:48}$ & 21:11 & $\bm{41 \; \mathrm{minutes}}$ & 1 hour 56 minutes \\
Oak & 21:10 & $\bm{21:30}$ & 20:45 & $\bm{20 \; \mathrm{minutes}}$ & 25 minutes \\
Mineral & 23:40 & $\bm{23:04}$ & 27:53 & $\bm{36 \; \mathrm{minutes}}$ & 4 hours 13 minutes \\
\hline
\end{tabular}
\end{center}
\end{table}

\section{Conclusion and outlook} \label{conclusion}

In this study a novel method for generating the early-stage fire arrival time of a wildfire based on active fire satellite detections obtained from polar-orbiting satellites is developed, implemented, and tested. The method treats the satellite measurements and the desired arrival times as random vectors and solves the problem of sampling from the distribution of the arrival times conditioned on an instance of the measurement. It accomplishes this using a conditional Wasserstein Generative Adversarial Network (cWGAN). Once the arrival time is inferred it can be employed in a coupled wildfire-weather prediction code to spin-up the atmosphere to an accurate initial state in order to make further predictions. 

There are two significant desirable features of the proposed method. First, by treating the inference problem in a probabilistic way, it generates an ensemble of likely fire arrival times. These can be used to generate a mean arrival time field - which serves as the best guess. They can also be used to compute the point-wise standard deviation in the arrival time which in turn can be used to quantify the uncertainty in the arrival time prediction and therefore in the initial state of the atmosphere. Second, the method is trained with simulations generated by a coupled atmosphere-wildfire prediction model. These simulations are consistent with the physics encoded in this code and through these the generator of the cWGAN learns to produce samples of arrival times that are also close to the underlying physics. 

When applied to data from four naturally occurring wildfires the cWGAN method yielded predicted fire perimeters that were more accurate than a similar method based on the SVM. In particular, a quantitative comparison revealed that the fire perimeter predictions produced by the cWGAN method were more accurate than the SVM predictions when measured using S\o rensen's coefficient (SC) and false alarm ratio (FAR). For the cWGAN, the average SC was 0.81 and the average FAR was 0.23, whereas for the SVM these values were 0.78 and 0.31, respectively. The SVM method performed better on the probability of detection (POD), with an average value of 0.92 compared with 0.87 for the cWGAN. The cWGAN method was also more accurate in predicting the ignition time of a fire, with an average error of 32 minutes, when compared with the SVM method which incurred an error of 2 hours and 11 minutes across the fires considered.

There are several avenues for future work through which the proposed method could be further improved. These include (a) conditioning the predictions of fire arrival time on important physical quantities like terrain and fuel maps; (b) extending this approach to solve the data assimilation problem of correcting the current state of an ongoing wildfire with newly acquired measurements; (c) considering the use of other conditional generative algorithms that avoid the pitfalls associated with adversarial training, like those based on diffusion maps \citep{song2020score}, for solving this problem.

\clearpage
\acknowledgments

BS, DR, and AAO gratefully acknowledge support from ARO, USA grant W911NF2010050. 
KH gratefully acknowledges support under NASA grant 80NSSC19K1091.
VC acknowledges support from the Center for Undergraduate Research at Viterbi Engineering (CURVE) program. 
The authors acknowledge the Center for Advanced Research Computing (CARC) at the University of Southern California, USA for providing computing resources that have contributed to the research results reported within this publication.

 \bibliographystyle{ametsocV6}
 \bibliography{references}

\end{document}